\documentclass[11pt]{article}

\usepackage[preprint]{acl}    
\usepackage{times}
\usepackage{latexsym}
\usepackage[T1]{fontenc}
\usepackage[utf8]{inputenc}
\usepackage{microtype}
\usepackage{inconsolata}
\usepackage{amsmath}
\usepackage{booktabs}
\usepackage{array}
\usepackage{tikz}
\usetikzlibrary{arrows.meta,positioning,fit,shapes.geometric,shapes.multipart}
\newcommand{\repourl}{https://anonymous.4open.science/r/RideWay-368B/}

\newcommand{\metric}{\ensuremath{U}}

\newcommand{\toolpen}{\ensuremath{\lambda_{\mathrm{tool}}}}
\newcommand{\turnpen}{\ensuremath{\lambda_{\mathrm{turn}}}}

\title{RideWay: Benchmarking Efficient Task Completion for Tool-Using Language Agents}

\author{
  Qingnuan Han\thanks{Equal contribution.}, Boli Fang\thanks{Equal contribution. Corresponding Author.}, Mingzhi Hou, Claire Liu\\
  Didi Global\\
  \texttt{summerhan, bolifang, houmingzhi, liujiaqiclaire@didiglobal.com}
}

\begin{document}
\maketitle

\begin{abstract}
AI agents are usually evaluated by whether they complete a task. In interactive service settings, a successful agent 
can still frustrate users by asking repeated questions, performing redundant searches, or making avoidable revisions. 
We introduce RideWay, an efficiency-centered benchmark for ridehailing agents in a stateful tool-calling environment, 
together with Efficiency Utility, a success-gated metric that discounts successful trajectories for excess tool calls 
and user-facing turns relative to task-specific reference effort. Human paired preferences calibrate the relative penalties, 
reflecting an aggregate service-workflow trade-off: extra dialogue often creates visible friction, whereas extra tool 
use can sometimes verify constraints or preserve user intent. Across 58 tasks and 24 models, the fitted penalty for 
excess turns is about twice that for excess tool calls. On task-disjoint held-out preferences, Efficiency Utility achieves 
78.7\% accuracy overall: 90.6\% when trajectories differ in turns, but chance-level accuracy when they differ solely in tool 
calls --- the axis on which human annotators agree least. RideWay therefore makes interaction efficiency measurable alongside 
task success, while exposing the boundary of count-based tool-use evaluation.
\end{abstract}

\section{Introduction}

Tool-using agents should be evaluated not only by whether they complete a task, but also by 
\emph{how} they get there. In ridehailing, two agents may both book a valid trip: one confirms the 
constraints and finishes directly, while another reaches the same outcome after repeated clarification or 
avoidable revision, imposing more visible burden on the user. Yet additional action is not necessarily wasteful: an agent 
that checks travel time to several candidate pickup points before booking may make more tool calls than one 
that guesses, but is more likely to preserve the user's constraints. A success-only evaluator cannot tell these cases apart, 
and a raw action-count penalty would collapse useful verification with needless friction. Evaluating successful 
agents therefore requires a success-gated efficiency measure that separates user-facing turns from backend tool 
use and learns their relative costs from human preferences.

Existing interactive-agent benchmarks make task completion increasingly measurable but often 
assign the same outcome score to successful trajectories that differ substantially in how they serve the user
\citep{liu2024agentbench,yao2024taubench,he2025vitabench,mazaheri2026agentatlas}. Accordingly, outcome-only scores cannot measure the 
distinct aggregate burdens of user-facing dialogue and backend work, even when trajectories reach the same outcome.

These considerations motivate RideWay\footnote{Code, data, and reproduction scripts:
\url{\repourl}. Appendix~\ref{app:reproducibility} details
what is included and how to reproduce every reported number.}, an efficiency-centered benchmark and scoring protocol for
ridehailing agents. Ridehailing is a strong testbed: interactions are naturally multi-turn and
tool-heavy, requiring location grounding, preference tracking, and mid-course revision, yet remain
concrete enough to evaluate against behavioral rubrics and backend state. We study this setting in Chinese, a high-resource language
that remains comparatively underrepresented in service-agent benchmarks. This setting raises a question that success-only evaluation leaves unanswered:

\emph{How do we measure the interaction efficiency of successful AI agents, not just whether they succeed?}

RideWay answers this question with Efficiency Utility, a success-gated score that assigns failures zero credit. For successful trajectories, 
it discounts only effort beyond task-specific reference floors \citep{anderson2018evaluation}, separately modeling excess user-facing turns 
and backend tool calls and calibrating their relative penalties from blinded human paired comparisons \citep{bradley1952rank}.

In summary, we contribute:
\begin{itemize}
    \item We introduce \textbf{RideWay}, a ridehailing evaluation
    benchmark for agents in a stateful tool-calling environment, with simulated users, panel-gated
    success judgments, and task-specific reference-effort annotations.
    \item We propose \textbf{Efficiency Utility}, a success-gated, reference-floor-relative metric
    that distinguishes excess tool calls from excess user-facing turns and fits their aggregate
    penalties from blinded human pairwise preferences.
    \item We validate Efficiency Utility on task-disjoint held-out human preferences and robustness checks. 
    Excess user-facing turns receive a consistently stronger, held-out-validated penalty, whereas lower agreement 
    on the tool axis identifies the boundary of what count-based efficiency can reliably resolve.
\end{itemize}

\section{Background and Related Work}
\label{sec:related-work}

Interactive-agent benchmarks have moved language-model evaluation from static answering to
multi-turn action in executable environments. Tool-use and reasoning-action work studies how
models interleave language with external actions or API calls
\citep{yao2023react,schick2023toolformer}, while function-calling and environment benchmarks test
API selection, schema following, web interaction, coding, and multi-step task execution
\citep{qin2024toolllm,yan2024bfcl,shridhar2021alfworld,yao2022webshop,deng2023mind2web,zhou2024webarena,jimenez2024swebench,liu2024agentbench,zhang2024pybench,yoran2024assistantbench,drouin2024workarena,ma2024agentboard}.
These benchmarks make agent behavior more observable, but their primary scores usually ask whether
the agent reaches a correct final state or completes the requested task.

Tool-agent-user benchmarks are the closest setting for RideWay because they require agents to
manage both dialogue state and tool state. $\tau$-bench and $\tau^2$-Bench instantiate
conversational agents that interact with users and tools in realistic domains
\citep{yao2024taubench,barres2025tau2bench}; VitaBench formalizes life-service tasks through user
state, database state, dialogue actions, tool invocations, and task rewards
\citep{he2025vitabench}; and T1-Bench extends this line to multi-scenario agent evaluation with
fixed simulator and judge configurations \citep{winata2026t1bench}. All three score task
completion or reward; none scores \emph{how} a successful trajectory got there. RideWay follows
their tool-agent-user premise but adds a second, orthogonal measurement layer on top of success:
a two-axis, human-calibrated effort penalty that separates backend tool calls from user-facing
turns rather than collapsing trajectory quality into a single completion signal.

Trajectory-sensitive evaluation studies the path an agent takes, not only its final answer. In
embodied navigation, Success weighted by Path Length gives credit only to successful agents and
discounts unnecessarily long paths \citep{anderson2018evaluation}. Code and agent benchmarks use
repeated-attempt or pass@k-style metrics to distinguish occasional success from reliable task
completion \citep{chen2021evaluating}. For service-style language agents, however, trajectory
effort is not a single path length: clarification turns, database checks, state-changing calls, and
preference-preserving revisions may affect the user differently. RideWay therefore measures two
observable effort axes, user-facing turns and backend tool calls, relative to task-specific
reference floors.

Human preference modeling provides a way to calibrate these effort axes when absolute scoring is
hard. Bradley--Terry paired-comparison models estimate latent preferences from comparative
judgments \citep{bradley1952rank}, and preference-based learning more broadly infers utility from
human choices rather than hand-specified scalar labels
\citep{ng2000algorithms,abbeel2004apprenticeship,christiano2017deep,ouyang2022training,rafailov2023direct,azar2024understanding,nika2024reward,zheng2023judging}. RideWay
uses this idea narrowly: annotators compare successful same-task trajectories for efficiency, and
the fitted penalties convert those judgments into a task-relative score. This differs from
efficient-LLM work on model compression, inference cost, latency, or carbon footprint
\citep{chen2023frugalgpt,wan2024efficientllms,zhu2024surveycompression,xia2024speculative,faiz2024llmcarbon},
which measures the computational system, not the user-facing interaction path.

\section{RideWay Benchmark}
\label{sec:benchmark}

\begin{figure*}[t]
\centering
\includegraphics[width=\textwidth]{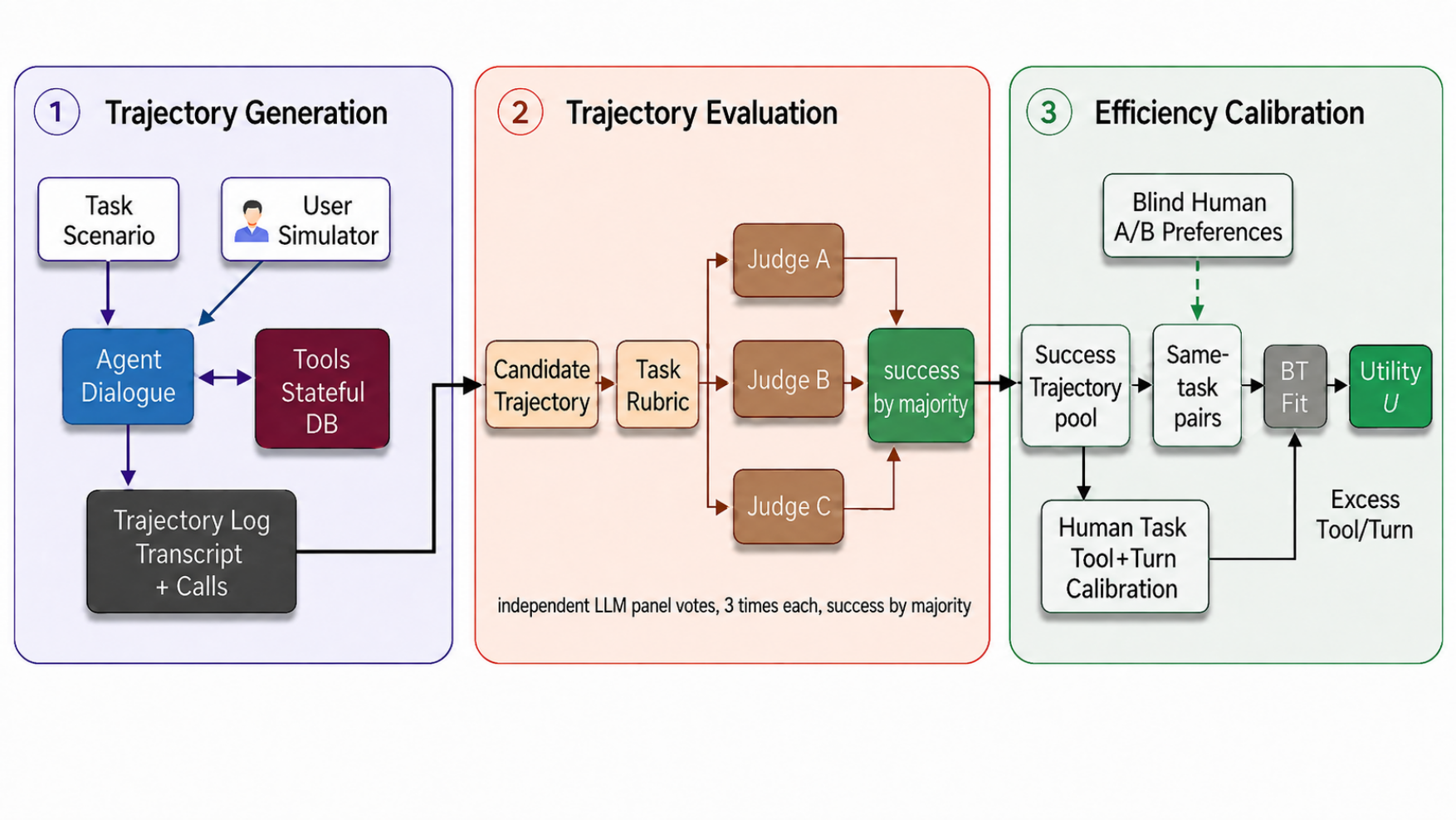}
\caption{RideWay's three-stage pipeline: trajectory generation, trajectory evaluation, and
efficiency calibration.}
\label{fig:framework}
\end{figure*}

RideWay evaluates ridehailing agents by generating stateful tool-agent-user trajectories and
gating each with an independent LLM success panel (Figure~\ref{fig:framework}). Each task is an
interactive episode between an evaluated agent, a simulated user, and a database-backed tool
environment, following recent tool-agent-user benchmarks
\citep{yao2024taubench,barres2025tau2bench,he2025vitabench,winata2026t1bench}: the agent converses
and calls ride-service tools until it completes the workflow or stops, producing a trajectory of
both user-facing dialogue and backend actions. A three-judge LLM panel then evaluates the trajectory
against the task rubric and backend state, so incomplete or constraint-violating attempts remain
visible as failures rather than efficient completions.

\paragraph{Design goals.}
The task suite in RideWay was constructed to make efficiency meaningful rather than incidental.
Tasks require multiple decisions, have more than one plausible solution path, can be graded with
rubrics and backend state, and include task-specific reference effort so a workflow with more
required constraints is not penalized merely for needing more interaction than a simpler one.

\begin{table}[t]
\centering
\footnotesize
\setlength{\tabcolsep}{4pt}
\renewcommand{\arraystretch}{1.05}
\begin{tabular}{@{}l
  >{\raggedleft\arraybackslash}p{1.6cm}
  >{\raggedleft\arraybackslash}p{1.6cm}
  >{\raggedleft\arraybackslash}p{1.6cm}@{}}
\toprule
Property & Aggregate & \texttt{rh44} & \texttt{custom14} \\
\midrule
Tasks & 58 & 44 & 14 \\
Reference tools & 3--22 (7.5) & 3--19 (6.9) & 5--22 (9.2) \\
Reference turns & 3--23 (9.1) & 3--22 (7.6) & 5--23 (13.6) \\
Rubrics & 2--11 (5.2) & 3--11 (5.7) & 2--5 (3.6) \\
\bottomrule
\end{tabular}
\caption{RideWay benchmark statistics, aggregate and by split. Numeric cells show min--max (mean)
per task; tool inventory and language are detailed below.}
\label{tab:benchmark-stats}
\end{table}

\paragraph{Environment and tools.}
The environment exposes 26 ride-service tools:
18 read tools and 8 write tools. Read tools cover
saved address lookup, current location, POI and nearby search, geocoding, history/order lookup,
ride estimates, trip-duration estimates, driver matching, flight information, route search, and
real-time ride status. Write tools cover car-type selection, immediate and scheduled ride
creation, ride confirmation, cancellation, destination updates, ride reminders, and ride notes.
The underlying state contains user profile fields, saved addresses, car types, dynamically
generated driver candidates, current and historical ride orders, POI data, route estimates, and
booking state. Tool calls can read or mutate this state, so agents must track both the user's latent
intent and the evolving environment.

\paragraph{Task construction.}
RideWay contains 58 manually curated ridehailing tasks (44-task \texttt{rh44} for calibration,
task-disjoint 14-task \texttt{custom14} for held-out validation). Tasks are organized around
capability tags that stress origin--destination grounding, order-type selection, explicit and
inferred preferences, context carryover, temporal reasoning, history recall, ambiguous locations,
driver recommendation, weak-signal inference, no-unsolicited-order behavior, route planning, and
mid-course revision. Many withhold constraints from the initial utterance or require preserving
earlier preferences after later edits.

\paragraph{Language and locale.}
RideWay is a Chinese-language benchmark: task instructions, simulated user utterances, and
user-facing agent responses use Simplified Chinese, while structured tool calls retain English API
identifiers, yielding a mixed Chinese/English interface. Its language- and culture-dependent
efficiency preferences should not be assumed to transfer unchanged to other locales.

\paragraph{Rubrics and reference effort.}
Each task includes a user scenario, an initial query, grading rubrics, and two human
reference-effort fields. To obtain the floors, Chinese-proficient annotators act as support agents
for each task, interacting with the same user LLM under the same instructions used to construct it;
we use the median of their tool-call and assistant-turn counts as $f_{\mathrm{tool}}(x)$ and
$f_{\mathrm{turn}}(x)$ (across 58 tasks: 3--22 calls, mean 7.5; 3--23 turns, mean 9.1; 2--11 rubric
items, mean 5.2). These floors are task-relative anchors, not theoretical shortest paths, used to
avoid comparing raw interaction length across different ride workflows.

\paragraph{LLM Panel Evaluation.} 
RideWay uses a conservative LLM-as-a-judge success criterion to avoid treating a lucky single
evaluation as solved \citep{zheng2023judging,panickssery2024llm}. Each trajectory is judged by a
panel of three distinct LLMs; each judge runs 3 times and takes a majority vote, and a trajectory
succeeds only if at least two of three judges pass it. This hierarchical aggregation reduces
reliance on any single model family's rubric interpretation. Judges run at temperature 0; the 3
passes still differ on 2--3\% of trajectories from genuine API-level non-determinism, so the
majority controls real noise rather than repeating an identical call. None of the three judges
(\texttt{hy3-preview}, \texttt{seed2.0-lite}, \texttt{grok4.3}) is in the 24-model agent pool,
limiting the self-preference bias that same-family judging can introduce \citep{panickssery2024llm}.

\section{Efficiency Utility Metric}
\label{sec:metric}

Efficiency Utility inherits its skeleton from Success weighted by Path Length
\citep{anderson2018evaluation}: a success gate, a task-relative reference floor, and a penalty only
for effort beyond that floor. Two changes adapt it from embodied navigation to service dialogue.
First, a single path length is the wrong unit: service-agent effort is user-facing (turns) and
backend (tool calls) at once, so the metric tracks two excess counts. Second, the relative cost of
these axes is not knowable a priori -- there is no ground-truth shortest interaction -- so the
trade-off is calibrated from human preferences (Section~\ref{sec:results}) rather than hand-set. The
exponential form in Eq.~\ref{eq:metric} then follows rather than being stylistic: fitting a
Bradley--Terry model requires an additively composable per-trajectory cost, and $\exp(-\cdot)$ is
the unique monotonic map turning such an additive cost into a per-unit multiplicative discount,
letting each excess call and turn contribute an independent factor.

Let $x$ be a task and $\tau$ be an agent trajectory for that task. Let
$c_{\mathrm{tool}}(\tau)$ and $c_{\mathrm{turn}}(\tau)$ denote its observed tool-call and dialogue-turn
counts. Tool calls count all structured tool invocations emitted by the assistant. Dialogue turns
count assistant messages sent to the user, excluding structured tool-call messages and terminal
control markers. For each task, the human-support annotation protocol in
Section~\ref{sec:benchmark} provides reference-effort floors $f_{\mathrm{tool}}(x)$ and
$f_{\mathrm{turn}}(x)$. These floors are task-relative anchors rather than statistics of the
evaluated model pool. We
define excess counts as
\begin{align*}
e_{\mathrm{tool}}(\tau) &= \max\{0, c_{\mathrm{tool}}(\tau)-f_{\mathrm{tool}}(x)\},\\
e_{\mathrm{turn}}(\tau) &= \max\{0, c_{\mathrm{turn}}(\tau)-f_{\mathrm{turn}}(x)\}.
\end{align*}
The \emph{Efficiency Utility} metric is thus defined as:
\begin{equation}
\metric(\tau) = \operatorname{success}(\tau)
\toolpen^{e_{\mathrm{tool}}(\tau)}
\turnpen^{e_{\mathrm{turn}}(\tau)},
\label{eq:metric}
\end{equation}
where $\operatorname{success}(\tau)$ indicates whether $\tau$ passes the 3-LLM panel
(Section~\ref{sec:benchmark}). A failed trajectory scores zero; a successful one at both floors
scores one; each additional tool call or turn multiplicatively reduces the score.

\section{Human Preference Calibration}

The metric's two penalties, $\lambda_{\mathrm{tool}}$ and $\lambda_{\mathrm{turn}}$, are estimated
in two steps: we construct same-task preference pairs from trajectories that already pass the
success gate -- so annotators judge only the efficiency of completed workflows -- and convert those
pairwise choices into the multiplicative penalties. This keeps success evaluation, reference-effort
annotation, and efficiency calibration separate while letting human preferences fix the trade-off
between the two observable effort axes.

\subsection{Constructing Preference Pairs}

Preference pairs are built after trajectory generation and judge-panel success filtering. For each
RideWay task, we collect model trajectories that pass the final panel gate and form pairs within
that successful same-task pool. Each comparison therefore contains two completed workflows for the
same user request. This construction removes two confounders from the human label: annotators are
not deciding whether a trajectory solved the task, and they are not comparing an easy task with a
hard one. They instead answer the narrower calibration question needed by Eq.~\ref{eq:metric}: when
both trajectories succeed, which completion imposes less excess interaction burden?

The sampler makes the penalty trade-off identifiable rather than merely collecting obvious wins. We
sample round-robin across tasks to cover the full \texttt{rh44} split, and within each task include
tool-only, turn-only, and mixed trade-off pairs (one trajectory using more tools but fewer turns);
the mixed cases reveal whether humans tolerate backend checks or extra dialogue more.
Chinese-proficient annotators see task context and full transcripts but not model identities,
counts, or utility scores, and label which trajectory is more efficient, ties allowed. We
spot-check pools from low-success models, whose rare panel-passing outputs may reflect evaluator
noise.

The resulting data are thus a task-controlled set of successful-trajectory comparisons, not a
generic transcript ranking, deliberately collected to expose the trade-off between extra backend
work and extra user-facing interaction. Held-out comparisons use the same rules on task-disjoint
tasks, so validation tests whether the learned trade-off transfers beyond the calibration tasks.

\subsection{Estimating Utility Penalties}

The second step estimates the two penalties from human preferences rather than hand-setting them,
since the axes differ in meaning: an extra turn often makes the user repeat or wait, while an extra
tool call may verify availability or preserve a preference after a revision. Fitting only on
same-task pairs that already pass the success gate, each successful trajectory is represented by a
linear excess-effort cost,
\begin{equation*}
C(\tau)=a e_{\mathrm{tool}}(\tau)+b e_{\mathrm{turn}}(\tau),
\end{equation*}
where $a$ and $b$ are nonnegative log-penalties for one excess tool call and one excess
user-facing turn. Since excess is measured above task-specific reference floors, $C(\tau)$ does not
penalize naturally longer tasks; it summarizes excess beyond the reference effort. The
cost is not a semantic judge of the transcript, but a compact preference-calibrated model of
observable burden
\citep{ng2000algorithms,abbeel2004apprenticeship,christiano2017deep,ouyang2022training,rafailov2023direct}.

We fit this cost with a Bradley--Terry paired-comparison model \citep{bradley1952rank}. For a
labeled pair $(\tau_i,\tau_j)$, the model predicts
\begin{equation*}
p_{ij}=P(i \succ j)=\sigma\!\left(C(\tau_j)-C(\tau_i)\right),
\end{equation*}
where $\sigma(z)=1/(1+\exp(-z))$. If trajectory $i$ has lower fitted cost than trajectory $j$,
$p_{ij}$ increases. We estimate $a$ and $b$ by minimizing regularized binary cross-entropy against
the human label $y_{ij}$, with $y_{ij}=1$ if $i$ is preferred, $0$ if $j$ is preferred, and
$\frac{1}{2}$ for a tie. Non-negativity encodes that extra effort should not raise utility, but the
constraint turns out to be slack: the unconstrained optimum already places both penalties strictly
positive (Section~\ref{sec:results}). A small L2 regularizer uses $\gamma=0.3$ (distinct from
$\toolpen$/$\turnpen$), and sweeping $\gamma$ from 0 to 8 changes $a/b$ by only 0.005.

Finally, we convert the fitted costs back to Eq.~\ref{eq:metric}: $\toolpen=\exp(-a)$ and
$\turnpen=\exp(-b)$, so successful trajectories satisfy $\metric(\tau)=\exp(-C(\tau))$. The score
stays in $[0,1]$, while human preferences determine whether excess turns should be discounted more
strongly than excess tool calls.

\section{Experimental Setup}
\label{sec:evaluation-design}

Our experiments evaluate complete RideWay trajectories rather than isolated final answers
\citep{zhou2024webarena,zhang2024pybench,jimenez2024swebench,he2025vitabench}. We run 24 non-thinking LLM configurations as agents,
grouped into four capability tiers reported alongside the leaderboard
(Table~\ref{tab:model-leaderboard}): \texttt{A} frontier proprietary systems, \texttt{B} strong
proprietary and open-weight systems, \texttt{C} mid-strength models, and \texttt{D} smaller models.
All agents use the same prompts, tool schemas, task order, simulator
configuration, and three-attempt budget. We exclude hidden-reasoning variants because their private
deliberation budgets are not comparable to the visible tool calls and user-facing turns measured by
Efficiency Utility.

The 44-task \texttt{rh44} split calibrates the metric and the task-disjoint 14-task
\texttt{custom14} split validates it -- not a random subsample: \texttt{custom14}'s long-horizon,
persona-driven, chit-chat-distracted tasks differ from \texttt{rh44}'s scenario-derived patterns
(Table~\ref{tab:benchmark-stats}: more reference turns, 13.6 vs.\ 7.6, and tool calls, 9.2 vs.\ 6.9,
but fewer rubric items, 3.6 vs.\ 5.7), testing generalization across construction. With 24 agents
and three attempts, the splits yield 3,168 and 1,008 raw trajectories; same-task pairs sampled from
panel-passing trajectories give 320 calibration votes (\texttt{rh44}) and 80 held-out votes
(\texttt{custom14}). GPT-4.1 is the fixed Chinese user simulator, and the success panel uses
\texttt{hy3-preview}, \texttt{seed2.0-lite}, and \texttt{grok4.3}.

\section{Results}
\label{sec:results}

\subsection{Evaluation on Success-Only Metrics}

We report Success Rate (SR), Pass@3, and Pass$^3$ -- the fraction of trials passing the
judge-panel gate, tasks solved in at least one of three attempts, and tasks solved in all three
\citep{chen2021evaluating} -- over 58 tasks, 24 models, and three attempts per model. Per-task SR
spans \textbf{12.5\%--87.5\%}; the top six-model SR band succeeds on \textbf{82.9\%} of trials,
while middle bands show the largest Pass@3--Pass$^3$ reproducibility gaps (\textbf{0.331--0.385};
Appendix~\ref{app:task-difficulty}). Per-model SR ranges from \texttt{mimo} (\textbf{87.4\%}) to \texttt{llama318b} (\textbf{2.3\%})
(Appendix~\ref{app:model-leaderboard}). These metrics are necessary success context, but they
collapse every successful trajectory to the same binary outcome.

\subsection{Why Success-Only Metrics Are Not Enough}
\label{sec:success-hides}

Table~\ref{tab:success-metric-miss-examples} shows why each successful trajectory needs an explicit trajectory-effort measurement.
Successful agents can have many excess tool calls with almost no extra dialogue, many excess turns with little extra tool use,
or excesses on both tool and turn axes. The final columns of Table~\ref{tab:success-metric-miss-examples} , computed with respect to human reference floor, show that success can 
come from different interaction paths.

Table~\ref{tab:excess-qualitative} describes the agentic inefficiencies in Table~\ref{tab:success-metric-miss-examples} with more concrete details. 
In the tool-excess example, agentic inefficiency comes from guessing points of interest before using
the provided destination; in the turn-excess example, it comes from the agent re-asking after delegation; and in the 
mixed-excess example, it comes from the agent re-asking, cancelling, and rebooking an already-started itinerary. In each case, the agent could
have acted on settled information instead of searching, asking, or rebooking again. This motivates
the calibration question: once success is known, how much should each excess
pattern cost?

\begin{table}[t]
\centering
\small
\setlength{\tabcolsep}{3pt}
\begin{tabular*}{\columnwidth}{@{\extracolsep{\fill}}llrrrr@{}}
\toprule
Task & Model & Tools & Turns & $e_{\mathrm{tool}}$ & $e_{\mathrm{turn}}$ \\
\midrule
\texttt{custom\_015} & \texttt{ministral3b} & 25 & 12 & \textbf{18} & 0 \\
\texttt{lc\_10} & \texttt{kimik26} & 7 & 18 & 1 & \textbf{7} \\
\texttt{mt\_09} & \texttt{qw37max} & 16 & 11 & \textbf{4} & \textbf{6} \\
\bottomrule
\end{tabular*}
\caption{Three successful trajectories spanning tool-only, turn-only, and mixed excess patterns;
These are the same trajectories detailed in
Table~\ref{tab:excess-qualitative}.}
\label{tab:success-metric-miss-examples}

\vspace{0.6em}
\footnotesize
\renewcommand{\arraystretch}{0.92}
\setlength{\tabcolsep}{3pt}
\begin{tabular*}{\columnwidth}{@{\extracolsep{\fill}}p{0.15\columnwidth}p{0.24\columnwidth}p{0.49\columnwidth}@{}}
\toprule
Excess & Goal & What the excess was \\
\midrule
Tool (18) & \texttt{custom\_015}: ride to a McDonald's & 5 wrong-guess POI searches before using
the given destination; retried bookings \\
Turn (7) & \texttt{lc\_10}: airport pickup & User twice said ``handle it''; agent re-asked the
same address 3$\times$ \\
Mixed (4, 6) & \texttt{mt\_09}: 3-stop errand + airport & Re-asked a flight number after being
told to decide; cancelled \& rebooked leg 1 \\
\bottomrule
\end{tabular*}
\caption{One real trajectory per excess pattern: avoidable redundancies over settled information.}
\label{tab:excess-qualitative}
\end{table}

\subsection{Efficiency Calibration and Pairwise Validation}

We evaluate Efficiency Utility on task-disjoint held-out pairwise preferences; the learned penalties are
asymmetric. In log-penalty form, the calibration estimates \textbf{\(a=0.1235\)} for tool excess and
\textbf{\(b=0.2448\)} for turn excess; equivalently,
\begin{equation*}
\mathbf{\lambda}_{\mathrm{tool}}=\mathbf{0.8838}, \qquad
\mathbf{\lambda}_{\mathrm{turn}}=\mathbf{0.7829}.
\end{equation*}
Thus one excess user-facing turn receives \textbf{2.01 times} the log-penalty of one excess tool
call (90\% bootstrap interval on \(b/a\): \([1.40,2.96]\), \(B=2500\)); the corresponding 90\%
bootstrap interval on \(a/b\) is \([0.34,0.71]\) (median 0.50).
Both penalties are strictly positive with \emph{disjoint} 90\% bootstrap intervals
(\(a\in[0.09,0.16]\), \(b\in[0.19,0.34]\)), and refitting without the non-negativity constraint
leaves them unchanged with \(a>0\) in every resample -- so the tool penalty is supported by the
data, not imposed by the constraint. (Appendix~\ref{app:reproducibility}) The fit is also stable to the reference floor: perturbing
every task's floor by $\pm1$--$2$ units and refitting moves held-out accuracy only within
$[0.787,0.813]$ and $\tau_b$ within $[0.573,0.627]$ (Appendix~\ref{app:human-annotation}).

This answers a natural objection: if some excess tool calls are useful verification, why does
Eq.~\ref{eq:metric} penalize every call? Because \(\lambda_{\mathrm{tool}}\) is fit from preferences
rather than assumed, it is not a per-call waste judgment but the population rate at which annotators
still preferred the higher-tool-call trajectory -- real but modest, penalizing tool excess about
half as hard per unit as turn excess. Table~\ref{tab:tool-axis-stats} shows the cost of this
per-call blindness: accuracy drops from \textbf{90.6\%} on pairs where turns differ to \textbf{chance
(50.0\%)} where tool calls are the only difference.

Table~\ref{tab:main-results} compares diagnostic variants: single-axis scores, equal-cost length
($e_{\mathrm{tool}}+e_{\mathrm{turn}}$), lexicographic rules that let one axis always dominate, and
monotone log/square-root variants. Coverage is the fraction of comparisons receiving a strict
ranking; accuracy is computed on directional answered pairs; and Kendall's $\tau_b$ is tie-aware
\citep{kendall1938new}.

\begin{table}[t]
\centering
\small
\setlength{\tabcolsep}{3pt}
\begin{tabular*}{\columnwidth}{@{\extracolsep{\fill}}lccc@{}}
\toprule
Metric & Coverage & Accuracy & $\tau_b$ \\
\midrule
Efficiency Utility & 1.00 & 0.787 & 0.573 \\
Tool-only & 0.95 & 0.394 & -0.200 \\
Turn-only & 0.71 & 0.925 & 0.600 \\
Equal-cost length & 1.00 & 0.773 & 0.547 \\
Turn-first lex. & 1.00 & 0.800 & 0.600 \\
Tool-first lex. & 1.00 & 0.427 & -0.147 \\
Log/sqrt forms & 1.00 & 0.787 & 0.573 \\
\bottomrule
\end{tabular*}
\caption{Held-out preference prediction for Efficiency Utility and diagnostic baselines, on
directional answered pairs. Efficiency Utility's 90\% bootstrap intervals: $[0.71,0.87]$ accuracy,
$[0.41,0.73]$ $\tau_b$ \citep{efron1979bootstrap}.}
\label{tab:main-results}
\end{table}

Efficiency Utility achieves \textbf{78.7\%} pairwise accuracy and \textbf{$\tau_b=0.573$} on
task-disjoint held-out preferences. Turn-only is accurate when it answers but misses tool-only
differences; tool-only and tool-first invert many mixed trade-offs where extra backend work reduces
dialogue. Turn-first lexicographic ordering has a marginally higher point estimate (0.800 vs.\
0.787 accuracy), but a paired bootstrap cannot separate the two (accuracy difference
$[-0.07,+0.04]$; $\tau_b$ difference $[-0.13,+0.08]$, see Appendix~\ref{app:reproducibility} for bootstrap details). We therefore treat them as statistically
tied and prefer Efficiency Utility because it remains a cardinal, averageable score rather than an
absolute-priority rule. The log/sqrt row is identical for pairwise ranking: any
strictly monotone transform preserves the comparison signs; Appendix~\ref{app:functional-form}
checks the cross-task averaging case where transforms can matter.

\paragraph{Where the tool axis stands.} Stratifying the 75 directional held-out pairs shows the
metric's strength and its limit (Table~\ref{tab:tool-axis-stats}). Efficiency Utility is accurate
when turns differ, but pure-tool pairs are at chance: the human vote splits exactly evenly between
fewer and more tool calls. The tool axis is therefore informative in aggregate but unreliable as a
per-call waste label: some extra tool calls are avoidable, while others preserve constraints or
check feasibility (Appendix~\ref{app:trajectory-examples}).
The strong 90.6\% turn-differing accuracy is also not uniform by margin: only \textbf{58.3\%}
($n=12$) on pairs differing by 1--3 excess turns, versus 100.0\% on the 41 pairs with a larger gap
(Table~\ref{tab:tool-axis-stats}) -- the headline number is carried disproportionately by
large-margin pairs.

\begin{table}[t]
\centering
\small
\setlength{\tabcolsep}{3pt}
\begin{tabular*}{\columnwidth}{@{\extracolsep{\fill}}lr@{}}
\toprule
Statistic & Value \\
\midrule
\metric{} accuracy, turn-differing pairs ($n=53$) & 90.6\% \\
\quad small turn diff ($|\Delta e_{\mathrm{turn}}|\le 3$, $n=12$) & \textbf{58.3\%} \\
\quad large turn diff ($|\Delta e_{\mathrm{turn}}|>3$, $n=41$) & 100.0\% \\
\metric{} accuracy, pure-tool pairs ($n=22$) & \textbf{50.0\%} \\
Human vote split, pure-tool pairs (fewer / more) & 11 / 11 \\
\bottomrule
\end{tabular*}
\caption{Held-out accuracy by turn-difference presence and margin size, and on pure-tool pairs
(with the human vote split).}
\label{tab:tool-axis-stats}
\end{table}

\subsection{Listwise Consistency Check}
\label{sec:listwise}

As a ranking-level consistency check on the same construct validated pairwise above, we test
whether Efficiency Utility can rank multiple successful solutions to the same
task, not only choose between two trajectories. We collect blind annotations on \textbf{10 held-out
tasks}: for each, the annotator places \textbf{8 successful trajectories} into four ordered
efficiency buckets, ties allowed, spanning both dual-axis and tool-dominant cases where turn-only
metrics cannot separate trajectories.

Efficiency Utility remains aligned with these judgments: task-level Kendall's $\tau_b$ is
positive on \textbf{all 10 tasks}, with mean \textbf{\(\tau_b=0.488\)}, median
\textbf{\(0.553\)}, and a 90\% bootstrap interval of \textbf{\([0.391,0.582]\)} (Appendix~\ref{app:reproducibility} for bootstrap details).
Tool-only is weaker (\(\tau_b=0.414\)) and negative on the hardest dual-axis task; turn-only ranks
only 8 of 10 tasks and is negative on 2 of those 8 -- both clearly worse than Efficiency Utility's
uniformly positive result. The listwise study thus supports the pairwise claim: the metric captures
the dominant cost of extra turns without collapsing tool-sensitive differences.

\subsection{Validation Boundaries: Human Ceiling and Judge Reliability}
\label{sec:iaa}

Two checks bound the results above rather than extending them: human agreement on the hardest
slice, and reliability of the success gate. For tool-axis IAA, a second blinded annotator relabels
the \textbf{23} held-out pairs that remain strictly pure-tool under the final floors. Human--human
agreement is \textbf{78.3\%} (Cohen's \(\kappa=0.658\) \citep{cohen1960coefficient}), higher than
Efficiency Utility's agreement with either annotator (\textbf{52.2\%}/\textbf{60.9\%},
\(\kappa=0.179/0.377\)); adding the metric as a third rater lowers Krippendorff's nominal
\(\alpha\) from \textbf{0.663} to \textbf{0.410}. By contrast, a 17-pair turn-differing relabel
check has \textbf{100.0\%} human agreement. The tool-axis boundary is therefore substantive: raw
tool counts cannot encode whether an extra call protected the user's goal or wasted effort.

\paragraph{Judge-panel reliability.}
\label{sec:judge-audit}
An annotator blind to model identity and to the panel verdict relabeled \textbf{75 trajectories}
success/fail using the same rubric, user scenario, and final system state given to the panel. Human
and panel agree on \textbf{90.7\%} of trajectories (Cohen's $\kappa=0.813$, 90\% CI
$[0.70,0.92]$, details at Appendix~\ref{app:reproducibility}). Agreement is perfect (30/30) for the twelve stronger models (Tier A/B, \S\ref{sec:evaluation-design});
all 7 disagreements (2 false positive, 5 false negative) come from the twelve weaker ones (Tier
C/D; Appendix~\ref{app:human-annotation}, Table~\ref{tab:judge-audit}). Most panel verdicts are unanimous (61/75, 3-0); the remaining 14/75
are 2-1 splits, and split verdicts are less reliable: 78.6\% human agreement versus 93.4\% for
unanimous cases.

\subsection{What Efficiency Reveals}
\label{sec:takeaway}

\begin{figure}[t]
\centering
\resizebox{\columnwidth}{!}{%
\begin{tikzpicture}[x=1cm, y=1cm, font=\scriptsize]
  \definecolor{tierA}{HTML}{2F5D8A}
  \definecolor{tierB}{HTML}{4E9C6F}
  \definecolor{tierC}{HTML}{C97A2B}
  \definecolor{tierD}{HTML}{9A9FA6}
  \tikzset{pt/.style={circle, draw=white, line width=0.35pt, fill=#1, minimum size=4.4pt, inner sep=0pt}}

  \draw[-, gray!35] (0,1.08) -- (7.6,1.08);
  \draw[-, gray!35] (0,2.16) -- (7.6,2.16);
  \draw[-, gray!35] (0,3.24) -- (7.6,3.24);
  \draw[-, gray!35] (0,4.32) -- (7.6,4.32);
  \draw[-, gray!35] (0,5.40) -- (7.6,5.40);

  \draw[-latex] (0,0) -- (7.9,0) node[right] {SR$_{58}$ (\%)};
  \draw[-latex] (0,0) -- (0,5.85) node[above] {U$_{\texttt{custom14}}$};
  \foreach \x/\xl in {0/0, 1.52/20, 3.04/40, 4.56/60, 6.08/80} \draw (\x,0) -- (\x,-0.08) node[below] {\xl};
  \foreach \y/\yl in {0/0.0, 1.08/0.2, 2.16/0.4, 3.24/0.6, 4.32/0.8, 5.40/1.0} \draw (0,\y) -- (-0.08,\y) node[left] {\yl};

  \node[pt=tierA] at (5.723,3.812) {};
  \node[pt=tierB] at (5.632,3.569) {};
  \node[pt=tierC] at (5.852,3.445) {};
  \node[pt=tierA] at (6.293,3.445) {};
  \node[pt=tierC] at (5.768,3.391) {};
  \node[pt=tierB] at (6.247,3.375) {};
  \node[pt=tierB] at (5.981,3.353) {};
  \node[pt=tierB] at (5.457,3.305) {};
  \node[pt=tierA] at (6.202,3.278) {};
  \node[pt=tierB] at (4.674,3.040) {};
  \node[pt=tierB] at (6.156,2.938) {};
  \node[pt=tierC] at (4.583,2.662) {};
  \node[pt=tierA] at (6.247,2.398) {};
  \node[pt=tierD] at (4.499,2.327) {};
  \node[pt=tierC] at (4.279,1.750) {};
  \node[pt=tierD] at (2.880,1.685) {};
  \node[pt=tierC] at (2.622,1.571) {};
  \node[pt=tierD] at (3.405,1.463) {};
  \node[pt=tierD] at (2.576,1.237) {};
  \node[pt=tierD] at (0.958,0.486) {};
  \node[pt=tierC] at (0.433,0.275) {};
  \node[pt=tierD] at (0.175,0.016) {};
  \node[pt=tierC] at (0.395,0.000) {};

  \node[circle, draw=black, line width=0.5pt, fill=tierB, minimum size=5.4pt, inner sep=0pt] (mimo) at (6.642,3.191) {};
  \node[align=left, anchor=south west, inner sep=1.5pt] (mimolabel) at (5.35,4.55) {\texttt{mimo}: highest SR,\\[-1pt]rank 10 on U};
  \draw[-latex, thin] (mimolabel.south) -- (mimo.north);

  \begin{scope}[shift={(4.75,0.15)}]
    \node[pt=tierA] at (0,1.35) {}; \node[right, xshift=2pt] at (0,1.35) {Tier A -- frontier proprietary};
    \node[pt=tierB] at (0,0.95) {}; \node[right, xshift=2pt] at (0,0.95) {Tier B -- strong proprietary/open};
    \node[pt=tierC] at (0,0.55) {}; \node[right, xshift=2pt] at (0,0.55) {Tier C -- mid-strength};
    \node[pt=tierD] at (0,0.15) {}; \node[right, xshift=2pt] at (0,0.15) {Tier D -- smaller models};
  \end{scope}
\end{tikzpicture}%
}
\caption{SR$_{58}$ vs. held-out \texttt{custom14} Efficiency Utility, all 24 models by tier.
Correlation is strong overall (\(\rho=0.785\)) but flattens among the top 12
(\(\rho=-0.161\)); \texttt{mimo} is the sharpest single case.}
\label{fig:sr-vs-u}
\end{figure}
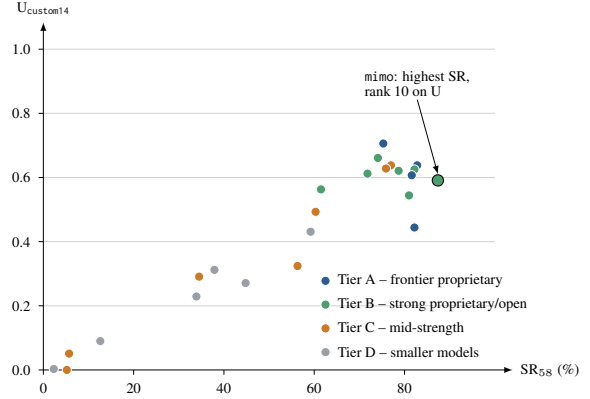

Having assessed Efficiency Utility's signal and limits, we ask what it shows. Figure~\ref{fig:sr-vs-u}
plots full-benchmark $\text{SR}_{58}$ against held-out \texttt{custom14} Efficiency Utility. We use
$\text{SR}_{58}$ rather than $\text{SR}_{\texttt{custom14}}$ here since it is the more reliable
signal (174 vs.\ 42 trials per model); the same-scope
$\text{SR}_{\texttt{custom14}}$-vs-$\text{U}_{\texttt{custom14}}$ relationship is reported
separately in Appendix~\ref{app:model-leaderboard}, where it is far more tightly coupled by
construction (Pearson $0.959$). The measures still
correlate overall (Spearman \(\mathbf{\rho=0.785}\), 90\% bootstrap interval \([0.521,0.933]\)), as expected
because failures score zero, but the relationship flattens among the 12 highest-U models
(\(\mathbf{\rho=-0.161}\), 90\% interval \([-0.664,0.420]\), details at Appendix~\ref{app:reproducibility}). The clearest case is \texttt{mimo}: it has the
highest SR of all 24 models (\textbf{87.4\%}) but ranks only \textbf{10th} on Efficiency Utility. Per-model
bootstrap intervals overlap between adjacent ranks, so the leaderboard is a coarse grouping rather
than a precise ordering; the robust point is that high success does not guarantee low burden.

The gap is not just a split artifact: \texttt{gpt55nothink} and \texttt{opus46} tie on
\texttt{custom14} success rate (\textbf{88.1\%}) yet differ by \textbf{11 points} in conditional efficiency on
successful trials (Appendix~\ref{app:model-leaderboard}). Because $\text{U}_{\texttt{custom14}}
\le \text{SR}_{\texttt{custom14}}$ by construction, this increment could in principle be vacuous;
it is not: across all 496 successful trajectories in the pairwise sets, only \textbf{28.8\%} have zero
excess tool calls and only \textbf{34.3\%} have zero excess turns (\textbf{10.3\%} have both at zero), so the
two-axis penalty is doing real discriminating work on most successful runs, not merely relabeling
success.

Among models with SR $>70\%$, excess interaction is model-specific: some trail on extra turns, while
others -- notably \texttt{mimo} and \texttt{gemini35flash} (SR $82\%$ but \metric{} rank 14) --
trail mainly on extra tool calls. Extra user-facing turns are the more reliable signal of human 
efficiency preferences, but both turns and tool calls capture model-dependent burden that success rates alone miss.

\section{Conclusion and Future Work}

RideWay closes a gap in service-agent evaluation: successful completion is necessary, but not
sufficient. Successful trajectories often differ in the excess interaction burden they impose, and Efficiency
Utility aligns with task-disjoint held-out pairwise human preferences, with consistent support from
a listwise ranking check and a clear tool-axis boundary: raw tool counts cannot encode the semantic usefulness of individual calls. 
Service-agent benchmarks should therefore reward agents not only for reaching task-level success, but also for reaching it with 
task-appropriate interaction effort.

Future work should test how far these findings and calibrated trade-offs transfer. RideWay is a Chinese ridehailing
benchmark, so the first step is broader validation across languages, domains, and real-user
settings. A second direction is richer semantic efficiency modeling: beyond RideWay's observable 
count-based axes, future metrics should distinguish wasteful tool calls from useful verification 
and incorporate latency, monetary cost, and user satisfaction. Finally, long-horizon tool workflows such as
agentic software engineering offer a natural testbed for studying whether task-relative efficiency
metrics generalize beyond ridehailing.

\clearpage

\section*{Limitations}
\label{sec:limitations}

RideWay is a single-domain, Chinese-language ridehailing benchmark. Its calibrated weights may
reflect local service norms about clarification, verification, and conversational length, and the
English tool identifiers may favor agents with stronger English API grounding. Cross-domain,
cross-language, and real-user validation are needed before treating the weights as general.

Efficiency Utility measures only observable turns and tool calls. It excludes hidden reasoning,
latency, monetary cost, tone, satisfaction, and safety, and raw counts simplify the distinction
between necessary work and waste. The held-out preference set is modest, excess turns are sparse,
and most preference labels come from one primary annotator; the IAA study is a contextual check,
not a replacement for broader labeling.

The benchmark also relies on automated components. Following VitaBench's configuration
\citep{he2025vitabench}, GPT-4.1 is the fixed user simulator, and success is gated by a three-judge
LLM panel (\texttt{hy3-preview}, \texttt{seed2.0-lite}, and \texttt{grok4.3}). These choices improve
reproducibility, and Section~\ref{sec:judge-audit} reports a direct human audit of the panel's own
reliability (90.7\% agreement, Cohen's $\kappa=0.813$ on a random sample of 75 trajectories), but
the audit does not cover every trajectory and does not replace real users or state-based
verification. Finally, the model pool intentionally excludes thinking models, and we do not report
token-cost correlations because tokenizer, verbosity, and serving differences make them
non-comparable across models.

\section*{Ethical considerations}
\label{sec:ethics}

The human labels in this study assess synthetic benchmark trajectories from a simulated
ridehailing environment rather than private customer data. The benchmark data do not contain
personally identifying information. Annotators were full-time interns or employees whose roles
include data annotation. They were informed that their labels would be used for benchmark
development and evaluation, and all annotation work was performed within their job duty.
Release artifacts exclude annotator identifiers, model identities in annotation interfaces, and
any tool/turn counts that would reveal the metric being validated. The codebase and trajectory
corpus were audited before release, and no offensive content was found. All human-facing tasks and
annotation instructions are in Simplified Chinese, with English tool identifiers shown as part of
the API interface. Appendix~\ref{app:human-annotation} describes the annotation protocol and its
limitations.

An efficiency metric can create pressure to reduce
interactions even when clarification would serve a user better. Efficiency Utility is therefore conditioned
on task success and should complement, not replace, evaluation of safety, correctness, user
control, and satisfaction. Because preferences about efficient service can vary across users and
domains, and because the present study is situated in Chinese-language ridehailing service,
future studies should test broader populations, languages, and task settings.

RideWay artifacts are intended for research evaluation of tool-using language agents, not for
deployment as a ridehailing service or for making operational decisions about real drivers,
passengers, or trips. The benchmark uses synthetic tasks, simulated users, and database states
constructed for evaluation. Released artifacts should be distributed with the MIT License and
documentation describing the language, domain, task splits, annotation protocol, and intended use.
The authors' institution does not maintain a separate ethics review board for this kind of study;
the work was conducted in accordance with recognized professional standards, including the ACM Code
of Ethics (\url{https://www.acm.org/code-of-ethics}).
The authors used AI writing and coding assistance during manuscript preparation and implementation;
all content, claims, and submitted text remain the responsibility of the authors.

\section*{Acknowledgements}

We thank Yueyue Han, Rui Li, Xiyue Zhang, Jixiang Guo, and Jiayun Peng for helping with the annotation work. 
We thank Claire Liu for discussions on benchmark design.

\clearpage

\bibliography{rideway_references}

\clearpage
\appendix

This appendix is organized as follows. Appendix~\ref{app:human-annotation} details the human
annotation protocol; Appendix~\ref{app:reproducibility} covers artifact release and reproduction;
Appendix~\ref{app:responsible-checklist} maps the paper to the Responsible NLP checklist;
Appendices~\ref{app:task-difficulty} and~\ref{app:model-leaderboard} report task-difficulty and
model-leaderboard details; Appendix~\ref{app:model-ids} lists model identifiers;
Appendix~\ref{app:functional-form} gives scoring and evaluation details; and
Appendix~\ref{app:trajectory-examples} reports detailed inefficiency-taxonomy 
examples.

\section{Human Annotation Protocol}
\label{app:human-annotation}

\subsection{Annotation tasks}

The human annotation in RideWay serves five measurement roles. First, human reference-effort
annotations provide task-specific floors for tool calls and assistant turns; the median count is
used as $f_{\mathrm{tool}}(x)$ and $f_{\mathrm{turn}}(x)$. Across the 58 tasks, the three
annotators' raw counts are close: the range (max$-$min) across the three annotators is on average
$1.97$ tool calls and $1.79$ turns (median $1.5$ and $2.0$), and only $5/58$ tasks have a tool-count
range of 5 or more (worst case: $[16,19,24]$), with just $1/58$ that wide on turns -- the median
floor is not typically riding on a single outlier annotator. We additionally checked whether the
calibration and its held-out validation depend on getting the floor exactly right: perturbing every
task's floor by the same $\delta \in \{-2,-1,+1,+2\}$ tool/turn units (clamped at 0) and refitting
$a,b$ from scratch on the perturbed training pairs changes held-out accuracy only within
$[0.787,0.813]$ and $\tau_b$(full) only within $[0.573,0.627]$, both at or above the
un-perturbed baseline ($0.787$, $0.573$) -- the metric is not brittle to plausible floor
mis-annotation in either direction. Second, the preference-calibration
and held-out validation sets ask for pairwise efficiency preferences between two successful
trajectories from the same task. Third, the listwise study asks an annotator to place eight
successful trajectories from one task into four ordered efficiency buckets. Fourth, the tool-axis
IAA study asks a second blinded annotator to relabel targeted held-out tool-axis pairs. Fifth, the
judge-panel audit (Section~\ref{sec:judge-audit}) asks a blind annotator to relabel a random
sample of trajectories success/fail, independent of and compared against the automated panel's own
verdict. The user simulator remains a fully automated component; the judge-panel success gate is
automated but is the one checked directly by this fifth role rather than assumed correct.

All natural-language task content, trajectories, and annotation instructions are presented in
Simplified Chinese. Structured tool names remain in English. The annotation task therefore
requires reading Chinese dialogue while interpreting English API identifiers.

\begin{table*}[t]
\centering
\small
\begin{tabular}{p{0.17\linewidth}p{0.20\linewidth}p{0.15\linewidth}p{0.37\linewidth}}
\toprule
Study & Annotation unit & Labels per unit & Recorded output \\
\midrule
Human reference effort & One interactive task episode & Team (variable per task) & Tool calls and assistant turns; task floor is the median count \\
Pairwise calibration/test & Two successful same-task trajectories & One primary vote & A more efficient, B more efficient, or tie \\
Listwise validation & Eight successful same-task trajectories & Ordered buckets & Four efficiency levels with ties allowed within a bucket \\
Tool-axis IAA & Targeted successful trajectory pair & Second blind vote & Independent A/B/tie judgment; primary label hidden \\
Judge-panel audit & One trajectory, any panel verdict & One blind vote & Success/fail judgment compared against the panel's own verdict \\
\bottomrule
\end{tabular}
\caption{Human annotation roles. ``Labels per unit'' describes the current experimental design,
not a recommendation that one primary label is sufficient for future benchmark versions.}
\label{tab:human-annotation-roles}
\end{table*}

\subsection{Construct and interface}

All human preference labels target the same narrow construct: efficiency as absence of avoidable
detours. The instructions explicitly exclude tone, wording, politeness, courtesy phrases, style,
and enthusiasm. The pairwise instruction states that both conversations complete the same task and
both succeed, then asks which side is more efficient, judging only whether the agent took detours
through redundant or repeated tool calls or avoidable conversational back-and-forth. Ties are an
allowed and encouraged answer when the trajectories are comparably efficient.

For reference-effort annotation, humans interact directly with the same user simulator used in
trajectory generation, acting as customer-service agents rather than passive judges. They have
access to the full ridehailing tool set and seek to complete the task by conversing with the
simulated user and calling tools as needed. We record each annotator's assistant-turn and tool-call
counts for the completed episode and use the median counts as the task's reference floors.

In the pair annotation interface, annotators see full transcripts and task context(task scenario and
grading rubric), but not model identities, tool-call counts,
turn counts, metric scores, or another annotator's decision. This is because annotators
sometimes need to know the task background so as to judge whether an apparent extra
tool call was necessary verification. For example, reapplying a driver preference after an order
edit should not be treated the same way as repeating a failed malformed call.

The listwise interface uses similar construct in an $n$-way format: for each task, eight
successful trajectories are sorted into four ordered buckets
(\emph{most efficient}, \emph{acceptable}, \emph{some waste}, \emph{very wasteful}), with ties
allowed within a bucket.

The judge-panel audit targets a different construct: task correctness rather than efficiency
preference. The annotator sees the same three inputs the judge panel's own prompt receives -- the
task's grading rubric, the user's original scenario, and the final system state (system time and
the ride order created, if any) -- alongside the full transcript, and gives a binary success/fail
judgment. Model identity, tier, and the panel's own verdict are hidden. Showing the rubric and
system state, not only the transcript, is deliberate: a fair check of the panel's reliability
requires the human referee to grade against the same standard and the same information the panel
used, not a looser or narrower one.

\subsection{Sampling and data-quality gates}

Efficiency comparisons are only constructed within the same task and only after all shown
trajectories pass the success gate. Candidate pairs and listwise sets are therefore not asking
whether one trajectory succeeded and another failed; they ask which successful path was cleaner.
Pair sampling is task-spanning rather than concentrated on a few high-contrast cases: pairs are
sampled round-robin across tasks, deduplicated against previous manifests, and mixed trade-off
pairs are included without forcing all pairs to be single-axis.

Three data-quality gates are applied before preference labeling. First, a candidate trajectory
must pass the judge-panel success gate. Second, candidate pairs are
deduplicated so the same trajectory pair is not relabeled under a new identifier. Third, newly
added low-success model trajectories are spot-checked before they enter the pair pool; this rule
was added after reward-passing but behaviorally unreliable trajectories from one low-success model
were retracted from pair construction.

The judge-panel audit is sampled differently by design: since its purpose is to check the success
gate itself, gating candidates on the panel's own verdict first would be circular. Its sample is
instead drawn uniformly at random from the full panel-output pool across all 24 pool models and
both domains, regardless of the panel's verdict.

\subsection{Current coverage and limitations}

The pairwise preference data are frozen at 320 panel-gated calibration votes over all 44
\texttt{rh44} tasks and 80 held-out votes over all 14 \texttt{custom14} tasks. The listwise study is finalized
at 10 tasks and 80 labeled trajectories; two remaining planned dual-axis tasks were not pursued
because the highest-priority tool-dominant coverage was complete. The reported tool-axis IAA
analysis covers 23 verified-pure-tool held-out pairs labeled by two annotators, plus a secondary
17-pair check on turn+mixed-axis pairs (Section~\ref{sec:iaa}). The judge-panel audit covers 75 randomly sampled
trajectories (Section~\ref{sec:judge-audit}), a subset of the full panel-output pool rather than
exhaustive coverage.

The main annotation limitation is that most pairwise items have one primary human vote. The IAA
study is designed to quantify this limitation rather than hide it. Its completed tool-axis result
shows moderate and imperfect human--human agreement on these targeted pairs
\citep{cohen1960coefficient}, which supports the paper's interpretation that tool-axis efficiency
has a lower human-agreement ceiling than a crisp correctness label would. The listwise study is
also a robustness check over the same construct, not an independent validation channel. The
judge-panel audit is likewise a partial check, not an exhaustive correctness audit of every
trajectory in the benchmark.

\subsection{Annotator recruitment, consent, and compensation}

Annotators were Chinese-proficient adults recruited by the authors for research annotation. They
were full-time interns or employees whose roles include data annotation, and all annotation work
was performed as part of their job duties during standard working hours; an hourly annotation rate
is therefore not applicable to this project. All annotators had university-level undergraduate or
graduate backgrounds; all were ages 18--35; 66.7\% were female and 33.3\% were male; they were
based in China and the United States. They were told that the annotations would be used to
construct and evaluate a benchmark, and they could skip unclear items by using the tie option when
neither trajectory was clearly more efficient. The annotation material contains synthetic
ridehailing dialogues and simulated tool states rather than private customer records. The
released artifact does not include annotator names, payment information, contact information, or
other direct identifiers.

\subsection{Detailed agreement statistics}

Table~\ref{tab:iaa-stats} gives the full tool-axis inter-annotator agreement statistics summarized
in Section~\ref{sec:iaa}, and Table~\ref{tab:judge-audit} gives the per-tier breakdown of the
judge-panel human audit summarized in Section~\ref{sec:judge-audit}.

\begin{table}[t]
\centering
\small
\setlength{\tabcolsep}{3pt}
\begin{tabular*}{\columnwidth}{@{\extracolsep{\fill}}lr@{}}
\toprule
Statistic ($n=23$ pairs) & Value \\
\midrule
A1--A2 raw agreement & 78.3\% \\
A1--A2 Cohen's $\kappa$ & 0.658 \\
\metric{}--A1 raw agreement & 52.2\% \\
\metric{}--A1 Cohen's $\kappa$ & 0.179 \\
\metric{}--A2 raw agreement & 60.9\% \\
\metric{}--A2 Cohen's $\kappa$ & 0.377 \\
Krippendorff's $\alpha$, humans only & 0.663 \\
Krippendorff's $\alpha$, humans + \metric{} & 0.410 \\
\bottomrule
\end{tabular*}
\caption{Tool-axis inter-annotator agreement statistics. A1/A2 are the two blinded annotators.
Efficiency Utility's agreement with either annotator sits below the human-human ceiling on every
statistic.}
\label{tab:iaa-stats}
\end{table}

\begin{table}[t]
\centering
\small
\setlength{\tabcolsep}{3pt}
\begin{tabular*}{\columnwidth}{@{\extracolsep{\fill}}lrr@{}}
\toprule
Tier & $n$ & Human--panel agreement \\
\midrule
A & 13 & 100.0\% \\
B & 17 & 100.0\% \\
C & 22 & 81.8\% \\
D & 23 & 87.0\% \\
\bottomrule
\end{tabular*}
\caption{Blind human relabel vs.\ the LLM judge panel's success/fail verdict, by model tier
($n=75$ randomly sampled trajectories). All 7 disagreements come from Tier C/D models.}
\label{tab:judge-audit}
\end{table}

\section{Reproducibility}
\label{app:reproducibility}

Code, data, and reproduction scripts are released at
\url{\repourl}.

\subsection{What is included}

The release contains the 58 RideWay task files with reference-effort annotations; the 320 training
and 80 held-out pairwise preference votes, each with the full trajectory pair embedded; the 80
listwise-tier labels over 10 tasks; the 23 verified-pure-tool-axis IAA pairs with both annotators'
votes; the 17
secondary turn+mixed-axis IAA pairs with both annotators' votes; the 75
trajectories relabeled for the human judge-panel audit (Section~\ref{sec:judge-audit}) with the
panel's own verdict alongside each human label; the 11 tool-axis discordant-pair taxonomy records
with classification; 15 hand-picked qualitative example trajectories; the released 24-model
leaderboard and 58-task difficulty ledger; and the scripts that compute every statistic in the
paper (Bradley--Terry fit, held-out baseline comparison, functional-form and lexicographic
ablations, IAA statistics, judge-panel audit statistics, listwise scoring, and the
leaderboard/difficulty-table generators). A top-level \texttt{HOW\_TO\_REPRODUCE.md} gives the
exact command for each reported number, its expected output, and the data-file schemas.

\subsection{How to run it}

Each script is self-contained given the small manifest/vote JSON files shipped alongside it (no
external corpus needed) and prints its result directly; \texttt{HOW\_TO\_REPRODUCE.md} lists the
exact invocation and expected output for every number in Sections~\ref{sec:metric}--\ref{sec:results}
and Appendices~\ref{app:task-difficulty}--\ref{app:model-leaderboard}. The two generator scripts
that regenerate the leaderboard and task-difficulty tables from raw trajectories are also included,
though they require the full simulation corpus rather than the shipped manifest files.

\subsection{Numbers requiring a non-default invocation}

A small number of robustness checks reported in the paper are not produced by the default,
no-argument invocation of a released script shown in \texttt{HOW\_TO\_REPRODUCE.md}'s headline
commands. Three are a documented flag or a small dedicated script already in the release; only the
first still requires an actual source edit, applied locally, since the script exposes no flag for
it:

\begin{itemize}
    \item \textbf{L2 sensitivity sweep} (Section~\ref{sec:metric}). The provided \texttt{fit()}
    routine takes an \texttt{l2} argument with default \texttt{0.3}. Calling it with
    \texttt{l2} $\in \{0, 2.0, 8.0\}$ instead and re-fitting reproduces the swept $a$, $b$, and held-out log-loss
    values.
    \item \textbf{Unconstrained Bradley--Terry fit} (Sections~\ref{sec:metric}
    and~\ref{sec:results}). The provided \texttt{fit()} routine clips both coefficients to stay
    non-negative after every gradient step; the analysis script's \texttt{--unconstrained} flag
    disables this clip for both the point fit and the bootstrap, reproducing the unconstrained
    point estimate and the confidence intervals on $a$ and $b$ reported in Section~\ref{sec:results}.
    \item \textbf{Turn-differing vs.\ pure-tool stratified accuracy} (Section~\ref{sec:results}). A
    dedicated script splits the held-out test pairs by whether the excess-turn difference is zero
    before scoring the \metric{} method, and separately tallies, on the zero-excess-turn subset, how
    often the human label matches the fewer-tool-calls side versus the more-tool-calls side,
    reproducing the two accuracy figures and the 11--11 split reported for the tool axis.
    \item \textbf{Floor-perturbation robustness} (Appendix~\ref{app:human-annotation}). A dedicated
    script perturbs every task's reference-effort floor by the same delta, recomputes excess tool
    calls and turns from the raw per-trajectory counts embedded in the calibration/held-out
    manifests, refits $a,b$ from scratch on the perturbed training pairs, and re-evaluates held-out
    accuracy and $\tau_b$ under that refit, reproducing the perturbation range reported in
    Appendix~\ref{app:human-annotation}.
\end{itemize}

\subsection{Bootstrap resample counts}

All reported 90\% intervals are percentile bootstrap intervals \citep{efron1979bootstrap}, used
uniformly across every analysis in the paper; most are read defensively (to show two quantities
cannot be separated), where a wider level would only reinforce the conclusion, rather than to
assert significance. The
pairwise held-out intervals (Table~\ref{tab:main-results}) use 10{,}000 resamples; the listwise
intervals (Section~\ref{sec:listwise}) use 5{,}000 resamples; the fit intervals on
$a$, $b$, $a/b$, and $b/a$ (Section~\ref{sec:results}) use 2{,}500 resamples (unchanged from an
earlier 500-resample pass, confirming that count was already stable); the judge-panel audit and IAA
control-set intervals (Section~\ref{sec:judge-audit}, Section~\ref{sec:iaa}) use 10{,}000
resamples; the model-level leaderboard intervals (Table~\ref{tab:model-leaderboard}) and the
functional-form ranking-equivalence interval (Appendix~\ref{app:functional-form}) use a task-level
bootstrap (resampling the 14 held-out tasks, not individual trials) at 5{,}000 resamples; the
SR-vs-U correlation intervals (Section~\ref{sec:takeaway}) resample the 24 (or 12) models directly
at 10{,}000 resamples.

\section{Responsible NLP Checklist Support}
\label{app:responsible-checklist}

This appendix maps the paper to the ARR Responsible NLP checklist and records checklist-specific
details that are not natural to include in the main text. The detailed protocols are intentionally
kept in the relevant appendices: Appendix~\ref{app:human-annotation} covers human annotation,
Appendix~\ref{app:reproducibility} covers artifact release and reproduction, and
Appendix~\ref{app:model-ids} lists evaluated model identifiers.

\paragraph{Task, data, and collection procedure.}
RideWay is a Chinese-language ridehailing benchmark for multi-turn tool-using agents. The task
suite, tool environment, user simulator, database state, task splits, rubrics, and success gate are
described in Sections~\ref{sec:benchmark}--\ref{sec:evaluation-design}. Data collection consists
of running 24 agent configurations for three attempts per task against a fixed simulated user and
database-backed ridehailing tool environment, then applying the LLM judge-panel success gate and
constructing successful same-task trajectory comparisons for efficiency annotation. Human
reference-effort floors are collected separately: annotators interact with the same user simulator
as customer-service agents, with access to the full tool set, and the median completed-episode
tool-call and assistant-turn counts define the task floors (Appendix~\ref{app:human-annotation}).

\paragraph{Human annotation protocol.}
Appendix~\ref{app:human-annotation} describes the annotation units, interfaces, instructions,
blinding, tie handling, sampling, data-quality gates, agreement statistics, and limitations. The
human annotation covers five roles: reference-effort floors, pairwise efficiency calibration and
held-out validation, listwise efficiency validation, tool-axis inter-annotator agreement, and a
human audit of the judge-panel success gate. Pairwise and listwise annotators see task context and
full transcripts but not model identities, tool/turn counts, metric scores, or another annotator's
decision; judge-panel audit labels are collected independently of the panel's verdict.

\paragraph{Annotator status and demographics.}
Appendix~\ref{app:human-annotation} is the source of truth for annotator recruitment, consent,
compensation status, and demographics. In brief, annotators were Chinese-proficient adult
full-time interns or employees whose roles include data annotation; all work was performed during
standard working hours as part of their job duties, so an hourly annotation rate is not applicable.
Appendix~\ref{app:human-annotation} also reports the collected education, age, gender, and
geographic information and states which annotator identifiers are excluded from the release.

\paragraph{Ethics review and professional standards.}
The authors' institution does not maintain a separate ethics review board for this kind of study.
The work was conducted in accordance with recognized professional standards, including the ACM Code
of Ethics (\url{https://www.acm.org/code-of-ethics}). The study uses adult annotators, synthetic
benchmark materials, and simulated tool states; it collects no sensitive personal data and releases
no annotator identifiers.

\paragraph{Privacy, content, and data protection.}
RideWay uses synthetic tasks, simulated users, and simulated database states rather than private
customer, passenger, driver, or trip records. The data do not contain personally identifying
information. Names, addresses, driver/order identifiers, preferences, and trip histories in the
released materials are synthetic benchmark fixtures. Annotation interfaces hide model identities,
tool/turn counts, metric scores, and other information that could reveal the metric being
validated. The released codebase and trajectory corpus were audited for offensive content, and no
offensive content was found.

\paragraph{Artifacts, license, and intended use.}
Code, data, and reproduction scripts are released at
\url{\repourl}. The release contains 58 synthetic
ridehailing tasks, 26 tools, 24 evaluated agent configurations, three attempts per task,
task-level reference-effort annotations, 320 calibration and 80 held-out pairwise preference
votes, 80 listwise labels, IAA labels, judge-panel audit labels, model leaderboards,
task-difficulty ledgers, and scripts for reproducing the reported statistics. In the
repository, the top-level \texttt{README.md} documents the benchmark, setup, and data/privacy
notes; \texttt{DATA\_CARD.md} documents provenance, intended use, limitations, and reporting
guidance; and \path{efficiency_utility_release/HOW_TO_REPRODUCE.md} documents the
Efficiency Utility artifacts, data schemas, commands, expected outputs, and reproduction workflow
(Appendix~\ref{app:reproducibility}). The
artifact is intended for research evaluation of tool-using language agents, not for deployment as a
ridehailing service or for operational decisions about real passengers, drivers, or trips. The
released artifact is distributed under the MIT License. External benchmarks, models, and methods
are cited in Sections~\ref{sec:related-work}--\ref{sec:evaluation-design}, and model identifiers
are listed in Appendix~\ref{app:model-ids}.

\paragraph{Computational experiments and reproducibility.}
Computational experiments are described in Section~\ref{sec:evaluation-design} and
Appendix~\ref{app:reproducibility}. The paper reports the model pool, task splits, trajectory
counts, preference-pair counts, judge-panel setup, regularization value, validation statistics,
bootstrap intervals, listwise ranking statistics, inter-annotator agreement statistics, and
judge-panel audit statistics. The study uses hosted LLM APIs rather than training new models, so
GPU-hour reporting for local model training is not applicable. Parameter counts and serving
infrastructure for proprietary hosted models are not available to the authors; open-weight model
identifiers and sizes, where known, are listed in Appendix~\ref{app:model-ids}.

\paragraph{Limitations, risks, and misuse.}
Limitations are discussed in the Limitations section. They include the single-domain,
Chinese-language ridehailing setting; reliance on simulated users and LLM judges; limited
held-out preference coverage; sparse excess-turn cases; one primary human vote for most pairwise
items; and the fact that Efficiency Utility measures observable turns and tool calls rather than
hidden reasoning, latency, monetary cost, tone, satisfaction, or safety. Risks and intended use are
discussed in the Ethical considerations section. In particular, optimizing for efficiency alone
could discourage useful clarification, so Efficiency Utility is conditioned on task success and
should complement, not replace, evaluation of correctness, safety, user control, and satisfaction.
RideWay should not be used as a ridehailing service, dispatch system, or source of factual travel
or safety guidance.

\paragraph{AI assistance.}
The authors used AI writing and coding assistance during manuscript preparation and
implementation. All submitted content, claims, analyses, and released artifacts remain the
responsibility of the authors.

\section{Task Difficulty Details}
\label{app:task-difficulty}

The task-difficulty analysis reports all 58 benchmark tasks with complete coverage: 24 models,
three agent attempts per model, and 72 trials per task. Table~\ref{tab:task-difficulty-extremes} gives the
hardest and easiest tasks by pooled task-level success rate. These repeated-attempt summaries are
reported in the spirit of pass@k-style evaluation \citep{chen2021evaluating}; they are task-level
summaries, not per-model leaderboards.

\begin{table*}[t]
\centering
\scriptsize
\setlength{\tabcolsep}{3pt}
\begin{tabular}{llrrr}
\toprule
Task & Domain & SR & Pass@3 & Pass$^3$ \\
\midrule
\multicolumn{5}{l}{\emph{Hardest tasks}} \\
\texttt{lc\_04\_cd\_taxi\_pref\_dismiss} & \texttt{rh44} & 12.5 & 0.250 & 0.042 \\
\texttt{custom\_028} & \texttt{custom14} & 18.1 & 0.292 & 0.083 \\
\texttt{custom\_029} & \texttt{custom14} & 18.1 & 0.292 & 0.083 \\
\texttt{mt\_08\_bj\_pet\_supermarket\_weekend} & \texttt{rh44} & 18.1 & 0.375 & 0.042 \\
\texttt{rp\_03\_cd\_furniture\_moving} & \texttt{rh44} & 23.6 & 0.417 & 0.125 \\
\midrule
\multicolumn{5}{l}{\emph{Easiest tasks}} \\
\texttt{mt\_03\_sh\_pickup\_two\_friends} & \texttt{rh44} & 80.6 & 0.875 & 0.708 \\
\texttt{mt\_02\_bj\_hotel\_pickup\_friend} & \texttt{rh44} & 83.3 & 0.917 & 0.792 \\
\texttt{custom\_015} & \texttt{custom14} & 84.7 & 0.875 & 0.792 \\
\texttt{custom\_030} & \texttt{custom14} & 84.7 & 0.875 & 0.792 \\
\texttt{airport\_new\_02\_xm\_current\_immediate} & \texttt{rh44} & 87.5 & 0.917 & 0.833 \\
\bottomrule
\end{tabular}
\caption{Task-level difficulty extremes. SR is the percentage of
successful individual agent attempts among 72 attempts. Pass@3 and Pass$^3$ are averaged over
models for three attempts per task.}
\label{tab:task-difficulty-extremes}
\end{table*}

Table~\ref{tab:sample-success-leaderboard} bands the 24 agents into four six-model reliability
tiers by SR, the summary referenced in Section~\ref{sec:results}.

\begin{table}[t]
\centering
\small
\setlength{\tabcolsep}{3pt}
\begin{tabular*}{\columnwidth}{@{\extracolsep{\fill}}lrrrr@{}}
\toprule
Reliability band & SR$_{58}$ & P@3$_{58}$ & P$^3_{58}$ & Gap \\
\midrule
Highest SR & \textbf{82.9} & \textbf{0.937} & \textbf{0.710} & 0.227 \\
Upper-mid SR & 75.5 & 0.911 & 0.580 & 0.331 \\
Lower-mid SR & 53.3 & 0.727 & 0.342 & \textbf{0.385} \\
Lowest SR & 15.7 & 0.267 & 0.058 & 0.209 \\
\bottomrule
\end{tabular*}
\caption{Model-level task-completion reliability bands on the 58-task RideWay benchmark. Models
are sorted by SR and averaged in four six-model bands. Gap is Pass@3 minus Pass$^3$.}
\label{tab:sample-success-leaderboard}
\end{table}

\section{Model Leaderboard Details}
\label{app:model-leaderboard}

Table~\ref{tab:model-leaderboard} reports the released 24-model evaluation summary. The table is
intended as a compact diagnostic view rather than the main validation result: SR, Pass@3, and
Pass$^3$ show task-completion reliability on the full 58-task suite, while held-out
\texttt{custom14} Efficiency Utility shows the efficiency-sensitive ordering under the final
preference-calibrated penalties. Models with similar SR can differ substantially in Efficiency
Utility when their successful trajectories require more excess turns or tools under the fitted
preference-calibrated penalties.

$\text{U}_{\texttt{custom14}}$ is the mean of $\metric(\tau)$ pooled directly over all 42 trials
(14 tasks $\times$ 3 attempts) in scope, not a per-task mean averaged again across tasks; failed
trials contribute $\metric(\tau)=0$ and are not excluded. Because it is based on only 14 held-out
tasks, Table~\ref{tab:model-leaderboard} reports a 90\% task-level bootstrap interval (5,000
resamples of the 14 tasks, pooling all trials of each resampled task) alongside every rank; the
released \texttt{leaderboard\_bootstrap\_ci.py} script reproduces every one of these 24 intervals
exactly from the raw trajectories, and also reports the stricter 95\% level. Every
adjacent-rank pair's interval overlaps -- Table~\ref{tab:model-leaderboard} should not be read as a
precise ordinal ranking. Meaningful separation instead requires a wide rank gap: rank 1
(\texttt{gpt55nothink}) is the first to separate from rank 14 onward, and rank 8 (\texttt{qw36p}) from
rank 16 onward. The table is more accurately read as identifying a clearly ahead top group, a
clearly behind bottom group, and a broad, statistically inseparable middle, rather than as
distinguishing any two neighboring models.

Because $\metric(\tau)=0$ on every failed trial, $\text{U}_{\texttt{custom14}}$ is capped by
$\text{SR}_{\texttt{custom14}}$: writing $\text{cond-U}$ for the mean of $\metric(\tau)$ over
successful trials only, $\text{U}_{\texttt{custom14}} = \text{SR}_{\texttt{custom14}} \times
\text{cond-U}$. Across the 24 models, $\text{SR}_{\texttt{custom14}}$ and
$\text{U}_{\texttt{custom14}}$ correlate strongly (Pearson $0.959$, Spearman $0.873$), so most of
the leaderboard's spread is success-rate-driven, and readers should not expect
$\text{U}_{\texttt{custom14}}$ to reorder models far from $\text{SR}_{\texttt{custom14}}$. The
increment is still real and visible in cond-U among models with comparable success rates: at
$\text{SR}_{\texttt{custom14}}=88.1\%$, \texttt{gpt55nothink} and \texttt{opus46} tie exactly, yet
their conditional efficiency differs by 11 points (cond-U $0.80$ vs.\ $0.69$) -- \texttt{opus46}'s
successful trajectories carry substantially more excess tool or turn effort. We report cond-U here
only as a diagnostic decomposition of $\text{U}_{\texttt{custom14}}$, not as an alternative ranking
score: unlike $\text{U}_{\texttt{custom14}}$, cond-U ignores success rate entirely and would reward
a model for quietly failing on its hardest attempts rather than completing them efficiently, so it
is not used for model comparison anywhere else in this paper. At the trajectory level rather than
the model level, the increment is broader than the aggregate correlation suggests: among all 496
distinct successful trajectories in the pairwise calibration and held-out sets, only $28.8\%$ have
zero excess tool calls and only $34.3\%$ have zero excess turns ($10.3\%$ have both at zero), so
the two-axis penalty is doing real discriminating work on the large majority of successful runs
even though its aggregate effect on the leaderboard ordering is modest.

\begin{table*}[t]
\centering
\scriptsize
\setlength{\tabcolsep}{3pt}
\begin{tabular}{rllrrrrrr}
\toprule
Rank & Model & Tier & SR$_{58}$ & Pass@3$_{58}$ & Pass$^3_{58}$ & SR$_{\texttt{custom14}}$ & U$_{\texttt{custom14}}$ & 90\% CI \\
\midrule
1 & \texttt{gpt55nothink} & A & 75.3 & 0.897 & 0.603 & 88.1 & 0.705 & [0.57, 0.83] \\
2 & \texttt{qw37max} & B & 74.1 & 0.879 & 0.586 & 92.9 & 0.661 & [0.56, 0.77] \\
3 & \texttt{qw3627b} & C & 77.0 & 0.914 & 0.603 & 78.6 & 0.638 & [0.48, 0.79] \\
4 & \texttt{sonnet5} & A & 82.8 & 0.931 & 0.724 & 78.6 & 0.637 & [0.49, 0.77] \\
5 & \texttt{gemma431b} & C & 75.9 & 0.914 & 0.586 & 73.8 & 0.628 & [0.48, 0.77] \\
6 & \texttt{qw37plus} & B & 82.2 & 0.931 & 0.690 & 83.3 & 0.624 & [0.50, 0.75] \\
7 & \texttt{glm52} & B & 78.7 & 0.931 & 0.638 & 81.0 & 0.620 & [0.50, 0.73] \\
8 & \texttt{qw36p} & B & 71.8 & 0.931 & 0.466 & 81.0 & 0.612 & [0.49, 0.73] \\
9 & \texttt{opus46} & A & 81.6 & 0.931 & 0.707 & 88.1 & 0.606 & [0.50, 0.71] \\
10 & \texttt{mimo} & B & 87.4 & 0.966 & 0.776 & 78.6 & 0.590 & [0.44, 0.73] \\
11 & \texttt{kimik26} & B & 61.5 & 0.759 & 0.466 & 73.8 & 0.563 & [0.41, 0.71] \\
12 & \texttt{dsv4pro} & B & 81.0 & 0.914 & 0.690 & 78.6 & 0.543 & [0.42, 0.66] \\
13 & \texttt{gemma426b} & C & 60.3 & 0.828 & 0.328 & 61.9 & 0.493 & [0.35, 0.64] \\
14 & \texttt{gemini35flash} & A & 82.2 & 0.948 & 0.672 & 76.2 & 0.443 & [0.33, 0.54] \\
15 & \texttt{qw359b} & D & 59.2 & 0.759 & 0.414 & 76.2 & 0.431 & [0.31, 0.55] \\
16 & \texttt{qw35ba3b} & C & 56.3 & 0.759 & 0.397 & 69.0 & 0.323 & [0.20, 0.46] \\
17 & \texttt{ministral8b} & D & 37.9 & 0.569 & 0.224 & 50.0 & 0.311 & [0.17, 0.46] \\
18 & \texttt{mistralsmall} & C & 34.5 & 0.569 & 0.155 & 40.5 & 0.291 & [0.16, 0.44] \\
19 & \texttt{ministral14b} & D & 44.8 & 0.690 & 0.224 & 47.6 & 0.270 & [0.13, 0.43] \\
20 & \texttt{ministral3b} & D & 33.9 & 0.534 & 0.121 & 38.1 & 0.229 & [0.12, 0.35] \\
21 & \texttt{granite418b} & D & 12.6 & 0.190 & 0.069 & 9.5 & 0.090 & [0.00, 0.22] \\
22 & \texttt{llama3370b} & C & 5.7 & 0.121 & 0.000 & 9.5 & 0.051 & [0.00, 0.14] \\
23 & \texttt{llama318b} & D & 2.3 & 0.069 & 0.000 & 2.4 & 0.003 & [0.00, 0.01] \\
24 & \texttt{nemotron3nano30b} & C & 5.2 & 0.121 & 0.000 & 4.8 & 0.000 & [0.00, 0.00] \\
\bottomrule
\end{tabular}
\caption{Released 24-model leaderboard summary. Models are sorted by held-out
\texttt{custom14} mean Efficiency Utility. SR, Pass@3, and Pass$^3$ are reported on the pooled
58-task suite under the final judge-panel success gate; SR$_{\texttt{custom14}}$ is success rate on
the same 14-task held-out split that U$_{\texttt{custom14}}$ is computed over, included so the
two columns can be read against each other directly (Appendix~\ref{app:model-leaderboard}
discusses this decomposition). The 90\% CI is a task-level bootstrap
interval (5,000 resamples of the 14 held-out tasks); every adjacent-rank pair's interval overlaps
(see the discussion above for how far apart two ranks must be to separate).}
\label{tab:model-leaderboard}
\end{table*}

Table~\ref{tab:model-leaderboard} reports SR only on the pooled 58-task suite and only
SR$_{\texttt{custom14}}$ (not Pass@3/Pass$^3$) on the held-out split, to keep the ranking table
narrow. Table~\ref{tab:model-leaderboard-full} reports the same three metrics separately on all
three scopes -- the pooled 58-task suite, the 44-task rh44 calibration split, and the 14-task
custom14 held-out split, plus U$_{\texttt{custom14}}$ alongside its own scope -- for readers who
want the full per-scope breakdown rather than the pooled/ranking view.

\begin{table*}[t]
\centering
\scriptsize
\setlength{\tabcolsep}{3pt}
\begin{tabular}{rlrrrrrrrrrr}
\toprule
Rank & Model & SR$_{58}$ & Pass@3$_{58}$ & Pass$^3_{58}$ & SR$_{\texttt{rh44}}$ & Pass@3$_{\texttt{rh44}}$ & Pass$^3_{\texttt{rh44}}$ & SR$_{\texttt{custom14}}$ & Pass@3$_{\texttt{custom14}}$ & Pass$^3_{\texttt{custom14}}$ & U$_{\texttt{custom14}}$ \\
\midrule
1 & \texttt{gpt55nothink} & 75.3 & 0.897 & 0.603 & 71.2 & 0.886 & 0.523 & 88.1 & 0.929 & 0.857 & 0.705 \\
2 & \texttt{qw37max} & 74.1 & 0.879 & 0.586 & 68.2 & 0.841 & 0.523 & 92.9 & 1.000 & 0.786 & 0.661 \\
3 & \texttt{qw3627b} & 77.0 & 0.914 & 0.603 & 76.5 & 0.932 & 0.568 & 78.6 & 0.857 & 0.714 & 0.638 \\
4 & \texttt{sonnet5} & 82.8 & 0.931 & 0.724 & 84.1 & 0.955 & 0.727 & 78.6 & 0.857 & 0.714 & 0.637 \\
5 & \texttt{gemma431b} & 75.9 & 0.914 & 0.586 & 76.5 & 0.932 & 0.591 & 73.8 & 0.857 & 0.571 & 0.628 \\
6 & \texttt{qw37plus} & 82.2 & 0.931 & 0.690 & 81.8 & 0.932 & 0.682 & 83.3 & 0.929 & 0.714 & 0.624 \\
7 & \texttt{glm52} & 78.7 & 0.931 & 0.638 & 78.0 & 0.932 & 0.636 & 81.0 & 0.929 & 0.643 & 0.620 \\
8 & \texttt{qw36p} & 71.8 & 0.931 & 0.466 & 68.9 & 0.932 & 0.409 & 81.0 & 0.929 & 0.643 & 0.612 \\
9 & \texttt{opus46} & 81.6 & 0.931 & 0.707 & 79.5 & 0.932 & 0.659 & 88.1 & 0.929 & 0.857 & 0.606 \\
10 & \texttt{mimo} & 87.4 & 0.966 & 0.776 & 90.2 & 0.977 & 0.795 & 78.6 & 0.929 & 0.714 & 0.590 \\
11 & \texttt{kimik26} & 61.5 & 0.759 & 0.466 & 57.6 & 0.727 & 0.409 & 73.8 & 0.857 & 0.643 & 0.563 \\
12 & \texttt{dsv4pro} & 81.0 & 0.914 & 0.690 & 81.8 & 0.909 & 0.705 & 78.6 & 0.929 & 0.643 & 0.543 \\
13 & \texttt{gemma426b} & 60.3 & 0.828 & 0.328 & 59.8 & 0.841 & 0.341 & 61.9 & 0.786 & 0.286 & 0.493 \\
14 & \texttt{gemini35flash} & 82.2 & 0.948 & 0.672 & 84.1 & 0.977 & 0.682 & 76.2 & 0.857 & 0.643 & 0.443 \\
15 & \texttt{qw359b} & 59.2 & 0.759 & 0.414 & 53.8 & 0.727 & 0.341 & 76.2 & 0.857 & 0.643 & 0.431 \\
16 & \texttt{qw35ba3b} & 56.3 & 0.759 & 0.397 & 52.3 & 0.727 & 0.364 & 69.0 & 0.857 & 0.500 & 0.323 \\
17 & \texttt{ministral8b} & 37.9 & 0.569 & 0.224 & 34.1 & 0.545 & 0.182 & 50.0 & 0.643 & 0.357 & 0.311 \\
18 & \texttt{mistralsmall} & 34.5 & 0.569 & 0.155 & 32.6 & 0.545 & 0.136 & 40.5 & 0.643 & 0.214 & 0.291 \\
19 & \texttt{ministral14b} & 44.8 & 0.690 & 0.224 & 43.9 & 0.727 & 0.182 & 47.6 & 0.571 & 0.357 & 0.270 \\
20 & \texttt{ministral3b} & 33.9 & 0.534 & 0.121 & 32.6 & 0.523 & 0.114 & 38.1 & 0.571 & 0.143 & 0.229 \\
21 & \texttt{granite418b} & 12.6 & 0.190 & 0.069 & 13.6 & 0.205 & 0.068 & 9.5 & 0.143 & 0.071 & 0.090 \\
22 & \texttt{llama3370b} & 5.7 & 0.121 & 0.000 & 4.5 & 0.114 & 0.000 & 9.5 & 0.143 & 0.000 & 0.051 \\
23 & \texttt{llama318b} & 2.3 & 0.069 & 0.000 & 2.3 & 0.068 & 0.000 & 2.4 & 0.071 & 0.000 & 0.003 \\
24 & \texttt{nemotron3nano30b} & 5.2 & 0.121 & 0.000 & 5.3 & 0.136 & 0.000 & 4.8 & 0.071 & 0.000 & 0.000 \\
\bottomrule
\end{tabular}
\caption{Full per-scope breakdown of Table~\ref{tab:model-leaderboard}: SR, Pass@3, and Pass$^3$
reported separately on the pooled 58-task suite, the 44-task rh44 calibration split, and the
14-task custom14 held-out split, under the final judge-panel success gate, plus
U$_{\texttt{custom14}}$ alongside its own scope. Rows are in the same
rank order as Table~\ref{tab:model-leaderboard} (sorted by held-out custom14 mean Efficiency
Utility).}
\label{tab:model-leaderboard-full}
\end{table*}

\section{Model Identifiers}
\label{app:model-ids}

Table~\ref{tab:model-ids} maps every shorthand used in this paper to the model family and version
it refers to, for the 24-model agent pool, the 3-judge success panel, and the fixed user simulator.

\begin{table*}[t]
\centering
\scriptsize
\setlength{\tabcolsep}{6pt}
\begin{tabular}{ll@{\hspace{18pt}}ll}
\toprule
Shorthand & Model & Shorthand & Model \\
\midrule
\texttt{opus46} & Claude Opus 4.6 & \texttt{gemma431b} & Gemma 4 31B \\
\texttt{sonnet5} & Claude Sonnet 5 & \texttt{gemma426b} & Gemma 4 26B \\
\texttt{gpt55nothink} & GPT-5.5 (non-thinking) & \texttt{qw3627b} & Qwen 3.6 27B \\
\texttt{gemini35flash} & Gemini 3.5 Flash & \texttt{qw35ba3b} & Qwen 3.6 35B-A3B \\
\texttt{qw37max} & Qwen 3.7 Max & \texttt{mistralsmall} & Mistral Small \\
\texttt{glm52} & GLM-5.2 & \texttt{llama3370b} & Llama 3.3 70B Instruct \\
\texttt{qw36p} & Qwen 3.6 Plus & \texttt{nemotron3nano30b} & Nemotron 3 Nano 30B-A3B \\
\texttt{qw37plus} & Qwen 3.7 Plus & \texttt{qw359b} & Qwen 3.5 9B \\
\texttt{kimik26} & Kimi K2.6 & \texttt{granite418b} & Granite 4.1 8B \\
\texttt{mimo} & MiMo v2.5 Pro & \texttt{ministral8b} & Ministral 8B \\
\texttt{dsv4pro} & DeepSeek V4 Pro & \texttt{ministral14b} & Ministral 14B \\
& & \texttt{ministral3b} & Ministral 3B \\
& & \texttt{llama318b} & Llama 3.1 8B Instruct \\
\midrule
\texttt{hy3-preview} & Hunyuan 3 Preview (judge) & \texttt{grok4.3} & Grok 4.3 (judge) \\
\texttt{seed2.0-lite} & Seed 2.0 Lite (judge) & GPT-4.1 & GPT-4.1 (user simulator) \\
\bottomrule
\end{tabular}
\caption{Shorthand-to-model mapping for all 24 pool models (grouped by Tier A/B/C/D, top to
bottom within each column, matching Table~\ref{tab:model-leaderboard}), the 3-judge success panel,
and the fixed user simulator.}
\label{tab:model-ids}
\end{table*}

\section{Scoring and Evaluation Details}
\label{app:functional-form}

For a successful trajectory, ranking by Efficiency Utility is equivalent to ranking by its log score,
\begin{equation*}
\log \metric(\tau) = e_{\mathrm{tool}}(\tau)\log\toolpen +
e_{\mathrm{turn}}(\tau)\log\turnpen.
\end{equation*}
Equivalently, it minimizes the fitted linear cost
$C(\tau)=a e_{\mathrm{tool}}(\tau)+b e_{\mathrm{turn}}(\tau)$, or after dividing by $a$,
$C_r(\tau)=e_{\mathrm{tool}}(\tau)+r e_{\mathrm{turn}}(\tau)$ with $r=b/a$. We report the
exponentiated form because it composes naturally with the success gate: failed trajectories receive
zero utility, reference-effort successful trajectories receive one, and excess interaction
multiplicatively discounts the score.
The tool-only and turn-only baselines set the other excess dimension to zero. The equal-cost
comparator instead ranks by $e_{\mathrm{tool}}+e_{\mathrm{turn}}$. Coverage is the fraction
of held-out comparisons on which a method provides a strict ranking; Kendall's $\tau_b$ is
tie-aware \citep{kendall1938new}. This distinction is essential because a method that ties nearly
all trajectories can have perfect conditional accuracy on the small subset it ranks.

\paragraph{Scope of the ranking-equivalence claim.} The log/sqrt row in
Table~\ref{tab:main-results} is bit-identical to Efficiency Utility's own row rather than merely
similar: any strictly monotonic concave transform of the per-trajectory cardinal score preserves
the sign of every pairwise comparison, so coverage, accuracy, and $\tau_b$ -- all pairwise,
within-task statistics -- are unchanged by construction. This equivalence is deliberately narrow: it
holds for ranking two trajectories against each other, but does \emph{not} extend to comparing the
\emph{mean} score of one trajectory set against another, because a monotonic transform does not
commute with averaging. The functional form therefore matters for cross-task aggregates such as
the mean-Efficiency-Utility column in Table~\ref{tab:model-leaderboard}, even though it is
invisible in Table~\ref{tab:main-results}. We use the exponential form for every reported aggregate
in this paper precisely because it is the one implied by the fitted Bradley--Terry cost (\S5.2), not
because the choice is inconsequential. We checked empirically whether this matters in practice by
recomputing every model's mean score with $1/(1+C(\tau))$ in place of $\exp(-C(\tau))$: the two
resulting orderings over all 24 models agree closely (Spearman $\rho=0.989$, 90\% task-level
bootstrap interval $[0.983,0.998]$), with the only
disagreement a 3-way reordering among \texttt{gemma431b}, \texttt{qw37plus}, and \texttt{opus46}
(ranks 5, 6, and 9 under the exponential form), whose scores under either form sit within 0.02
of each other. The functional-form choice is therefore consequential in principle but not
disruptive on this data.

The task-difficulty summary in Appendix~\ref{app:task-difficulty} is computed from the
58-task, 24-model trajectory set under the same judge-panel success gate used throughout the
paper. Pairwise calibration, held-out ablations, lexicographic baseline checks, listwise
validation, IAA, and taxonomy findings use the same filtered trajectory pools described in
Sections~\ref{sec:metric}--\ref{sec:results}. The judge-panel audit (Section~\ref{sec:judge-audit})
is the one exception: it deliberately samples from the \emph{unfiltered} panel-output pool,
including panel failures, since checking the gate itself requires not conditioning on the gate's
own decision. The tool-axis inter-annotator study is summarized in
Appendix~\ref{app:human-annotation}.

\section{Detailed Inefficiency-Taxonomy Examples}
\label{app:trajectory-examples}

The taxonomy inspects tool-axis held-out pairs where Efficiency Utility disagrees with the
human vote. For each discordant pair, the analysis focuses on the higher-tool trajectory that the
human preferred but the metric ranked lower. Table~\ref{tab:inefficiency-patterns} reports the two
representative worked cases, in the same format as Table~\ref{tab:excess-qualitative}: what looked
like avoidable excess by count was not.

\begin{table}[t]
\centering
\footnotesize
\renewcommand{\arraystretch}{0.92}
\setlength{\tabcolsep}{3pt}
\begin{tabular*}{\columnwidth}{@{\extracolsep{\fill}}p{0.29\columnwidth}p{0.21\columnwidth}p{0.42\columnwidth}@{}}
\toprule
Case & Goal & What the extra calls did \\
\midrule
Preference preservation (\texttt{c14test\_026}) & Re-apply a no-smoking, male-driver preference
after a time change & EU's pick (5 calls) drops the preference; the human-preferred trajectory (10
calls) re-matches and reapplies it on both bookings \\
Multi-stop verification (\texttt{c14test\_055}) & Plan a multi-segment sightseeing circuit &
EU's pick uses 20 calls; the human-preferred trajectory uses 28, with 18 (64\%) checking travel
time between stops before committing to a route order \\
\bottomrule
\end{tabular*}
\caption{Two tool-axis discordant pairs where the higher-tool trajectory only looked wasteful by
count: the extra calls preserved a preference or verified a route, not padding.}
\label{tab:inefficiency-patterns}
\end{table}

\paragraph{Takeaway.}
In the taxonomy, all 11 discordant tool-axis cases are verification-dominant; none is ambiguous and
none is a clean error-dominant case. The result supports the paper's conservative
interpretation: the weakest axis is not simply ``extra tool calls are waste,'' but rather that raw
tool counts cannot always separate unnecessary calls from useful verification, comparison, and
preference-preservation work.

\end{document}